%% file: main.tex
\documentclass[letterpaper, 10 pt, conference]{ieeeconf}  

\IEEEoverridecommandlockouts                              

\usepackage{graphicx}
\usepackage{listings}
\usepackage{float}
\usepackage{subcaption}
\usepackage{xcolor}
\usepackage{hyperref}
\usepackage{tabularx}
\hypersetup{
    colorlinks = true,
    linkbordercolor = blue
}

\title{\LARGE \bf
Talk Through It: End User Directed Manipulation Learning
}

\author{Carl Winge, Adam Imdieke, Bahaa Aldeeb, Dongyeop Kang, Karthik Desingh
\thanks{C. Winge, A. Imdieke, B. Aldeeb, D. Kang, and K. Desingh are with the Minnesota Robotics Institute, University of Minnesota, Twin Cities, Minneapolis, USA \{\tt winge134, imdie022, baldeeb, dongyeop,  kdesingh\}@umn.edu}
}

\begin{document}

\maketitle
\thispagestyle{empty}
\pagestyle{empty}

\begin{abstract}

Training generalist robot agents is an immensely difficult feat due to the requirement to perform a huge range of tasks in many different environments. We propose selectively training robots based on end-user preferences instead.

Given a \textit{factory model} that lets an end user instruct a robot to perform lower-level actions (e.g. `Move left'), we show that end users can collect demonstrations using language to train their \textit{home model} for higher-level tasks specific to their needs (e.g. `Open the top drawer and put the block inside'). We demonstrate this hierarchical robot learning framework on robot manipulation tasks using RLBench environments. Our method results in a 16\% improvement in skill success rates compared to a baseline method.

In further experiments, we explore the use of the large vision-language model (VLM), Bard, to automatically break down tasks into sequences of lower-level instructions, aiming to bypass end-user involvement. The VLM is unable to break tasks down to our lowest level, but does achieve good results breaking high-level tasks into mid-level skills. We have a supplemental video and additional results at \url{talk-through-it.github.io}.

\end{abstract}

\input{sections/00_introduction}
\input{sections/01_related}
\input{sections/02_methodology}
\input{sections/03_experimental_setup}
\input{sections/04_conclusion}


\addtolength{\textheight}{0cm}   
\section*{Acknowledgements}
This project is funded by the MnRI Seed Grant from the Minnesota Robotics Institute. We thank Chahyon Ku for insightful discussions and proof reading this paper.
\bibliographystyle{ieeetr}
\bibliography{references}
\newpage
\section*{Appendix}
\input{sections/appendix}

\end{document}

%% file: sections/00_introduction.tex
\section{INTRODUCTION}

Interactive devices such as Siri, Alexa, and Google Home have brought AI into the home, but the prospect of having embodied AI manipulating household objects remains out of reach. 
Home robots will require personalization and adaptation to their specific environments to carry out tasks. We believe that it will be the end users who actively imbue the domestic robots with skills and behaviors pertinent to the tasks they intend their robots to assist with, thus necessitating the development of robot learning frameworks centered around \textit{end users}. 

\begin{figure}[t]
\includegraphics[width=0.5\textwidth]{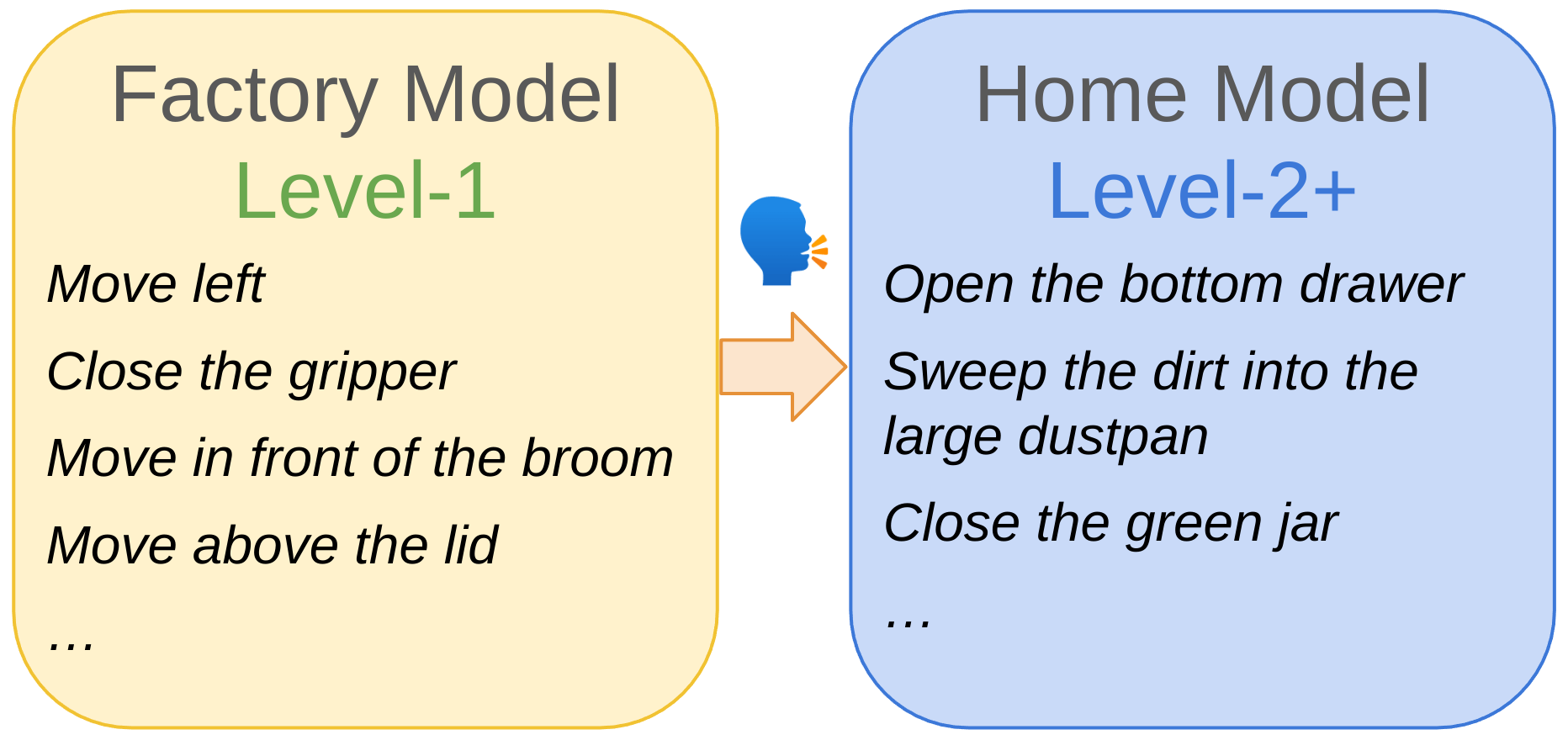}
  \caption{A level-1 \textit{factory model} is trained with a diverse set of primitive action commands. End users train the robot in their homes to complete the tasks they care about by using the level-1 commands to collect demonstrations. We call this the \textit{home model}.}
  \label{fig:overview}
\end{figure}

We involve end users in robot manipulation learning by decomposing it into two steps: creating a \textit{factory model} and creating a \textit{home model}, as shown in Figure~\ref{fig:overview}. 
We envision a user receiving a robot programmed with a \textit{factory model} which endows it with primitive capabilities, allowing it to follow basic instructions such as ``move right", ``close the gripper", or ``move above the green jar". The end user would bootstrap off of these capabilities to direct the robot's \textit{factory model} to evolve into a personalized \textit{home model}. By instructing the robot through more complex skills such as ``sweep the dust into the dustpan", or ``open the top drawer", the user can teach the robot to follow more complex instructions.

Learning from demonstration provides an accessible method of robot training through behavior cloning. Our work presents a framework to collect demonstrations by leveraging actions already learned by a robot agent. We show that we can train a robot to perform primitive actions and then use those actions to abstract away low-level control. Using natural language, we instruct our robot through skill and task demonstrations that we use to train the robot and expand its capabilities.

We include experiments using a large vision-language model (VLM) to see whether it can replace a human breaking down tasks for a robot. While the VLM can break a task into skills, it cannot break skills into primitive actions. We quantitatively show that end user directed robot learning is still necessary to close the domain gap between low-level reasoning (grounded on the current scene and task) and high-level reasoning (common knowledge).

Through this work, we present an end user directed hierarchical robot learning framework, leveraging natural language for communication. 
\begin{itemize}
    \item We present a method for training a robot to respond to observation-dependent and observation-independent language commands for primitive actions. 
    \item We show language commands for primitive actions can be used in sequence to collect demonstrations for training higher-level skills and tasks. This technique allows non-expert end users to train a robot according to their personal needs.
    \item Our results prove that our hierarchical training method leads to performance gains over a state-of-the-art baseline method which doesn't utilize hierarchical training.
    \item We demonstrate a VLM can successfully chain skills learned by our model to complete longer-horizon tasks.
\end{itemize}

%% file: sections/01_related.tex
\section{RELATED WORK}

Our work entails robot manipulation learning from end-user demonstrations in a hierarchical fashion. In this section, we list related work in the areas of learning from demonstrations, demonstration data acquisition, and reasoning via large vision-language models.

\subsection{Learning from Demonstrations}
Behavior cloning (BC) has garnered significant attention when it comes to robot manipulation task learning from demonstrations \cite{rt22023arxiv, jang2021bcz, shridhar2022peract, gervet2023act3d}. For our proposed framework, we require the policy to be both capable of learning various skills and sample efficient when it comes to training. BC models such as BC-Z~\cite{jang2021bcz} and MOO~\cite{moo2023arxiv} are designed to output robot end-effector states based on expert demonstrations. RT-2~\cite{rt22023arxiv} leveraged vision language models (VLMs) that generate text containing end-effector pose.  While such methods demonstrated impressive success rates, they required prohibitively large amounts of demonstration data. 

To address the data collection bottleneck, Wang et al. \cite{wang2023mimicplay} proposed using videos of human play to augment the training process. Recording the human play data is fast, but they still require data collection using robot teleoperation to transfer the skills to their robot. Shridhar et al. proposed PerAct~\cite{shridhar2022peract}, which demonstrated high accuracy and sample efficiency. The PerAct architecture is suitable for our purposes because it is sample efficient and leverages the RLBench~\cite{james2019rlbench} simulation environment, which includes many benchmark manipulation tasks.

\subsection{Demonstration Data Acquisition}
BC models require expert demonstrations. However, the tools used for interfacing with robots are not particularly user-friendly.
Teleoperation using joysticks or other controllers is commonly used to control the robot's end-effector \cite{wang2023mimicplay}, with some leveraging VR \cite{jang2021bcz, clever2021assistive} to facilitate the data collection process. 
ALOHA \cite{zhao2023learning} presented the idea of using a twin robot for the target robot to mimic. 
While these teleoperation techniques have become more intuitive, they still require expensive equipment that can be difficult to set up. Furthermore, a recent user study from Mahadevan et al.~\cite{mahadevan2022mimic} shows that it is easier for users to control a robot when they can command meaningful actions instead of directly controlling the robot motion. 
In our work, we focus on exposing the robot's controls to a non-expert end user via natural language. Specifically, the \textit{factory model} is trained on expert demonstrations, and \textit{home models} are trained using demonstration data collected via natural language.  

With the progress of large language models and large vision-language models, language is becoming the go-to method for task specification. Language is mostly used to expose the robot's learned skills to an end user \cite{shridhar2022peract, ahn2022i, driess2023palm}. Lynch et al.~\cite{lynch2022interactive} bridge the gap between teleoperation and task execution by allowing users to control the robot through speech in real-time. However, unlike our method, their methods do not use language to collect demonstrations. 

\subsection{Reasoning via Large Vision-Language Models: }
Pre-trained large language models (LLMs)~\cite{anil2023palm, brown2020language} have demonstrated impressive reasoning, prompting a new generation of LLM-based Vision-Language Models (VLMs) that extend LLMs to allow reasoning over visual contexts~\cite{driess2023palm, rt22023arxiv}. These VLMs demonstrate the ability to describe visual scenes and answer questions about them, as well as control robots~\cite{driess2023palm}. Given a prompt describing a situation and intention, these VLMs demonstrated the ability to reason over tasks \cite{driess2023palm}  and integrate feedback from their environments~\cite{ahn2022i, huang2022inner}. We utilize the publicly available Bard model~\cite{bard_google_ai} in our VLM experiments.

%% file: sections/02_methodology.tex
\section{Framework}
\subsection{Framework Architecture}
\begin{figure}
    \begin{subfigure}{\linewidth}
    \includegraphics[width=\linewidth]{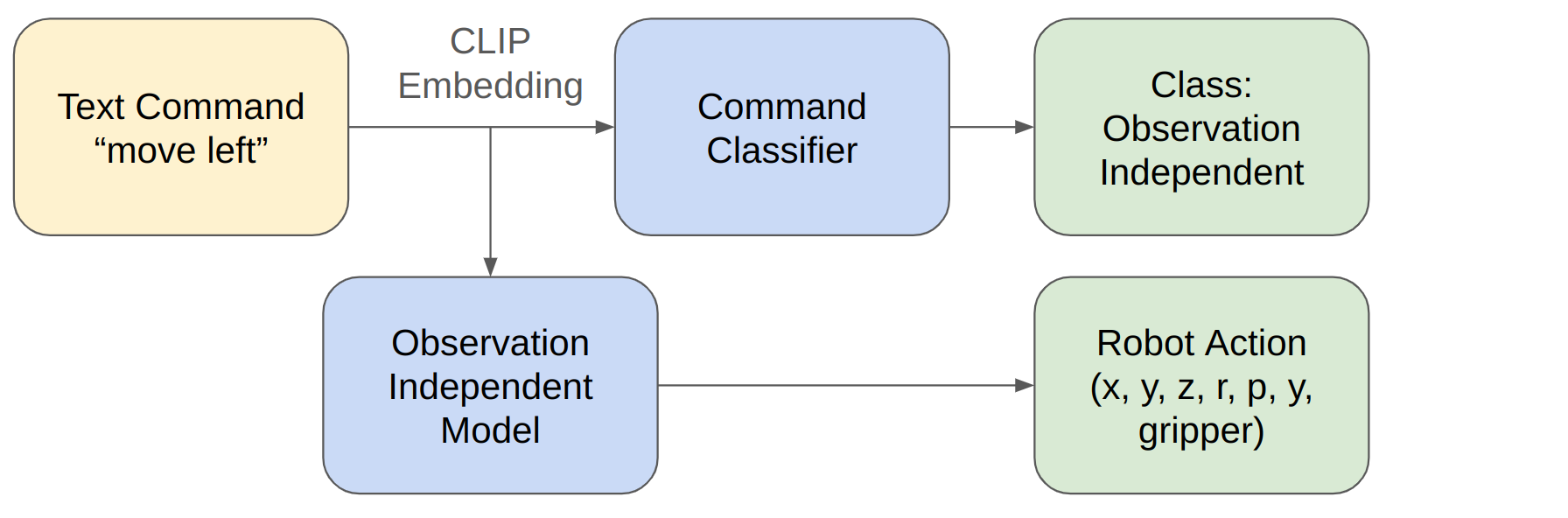}
    \caption{If the command classifier determines a command is observation-independent, the observation-independent model uses the text embedding to output a robot action, as shown above.}
    \end{subfigure}

    \begin{subfigure}{\linewidth}
    \includegraphics[width=\linewidth]{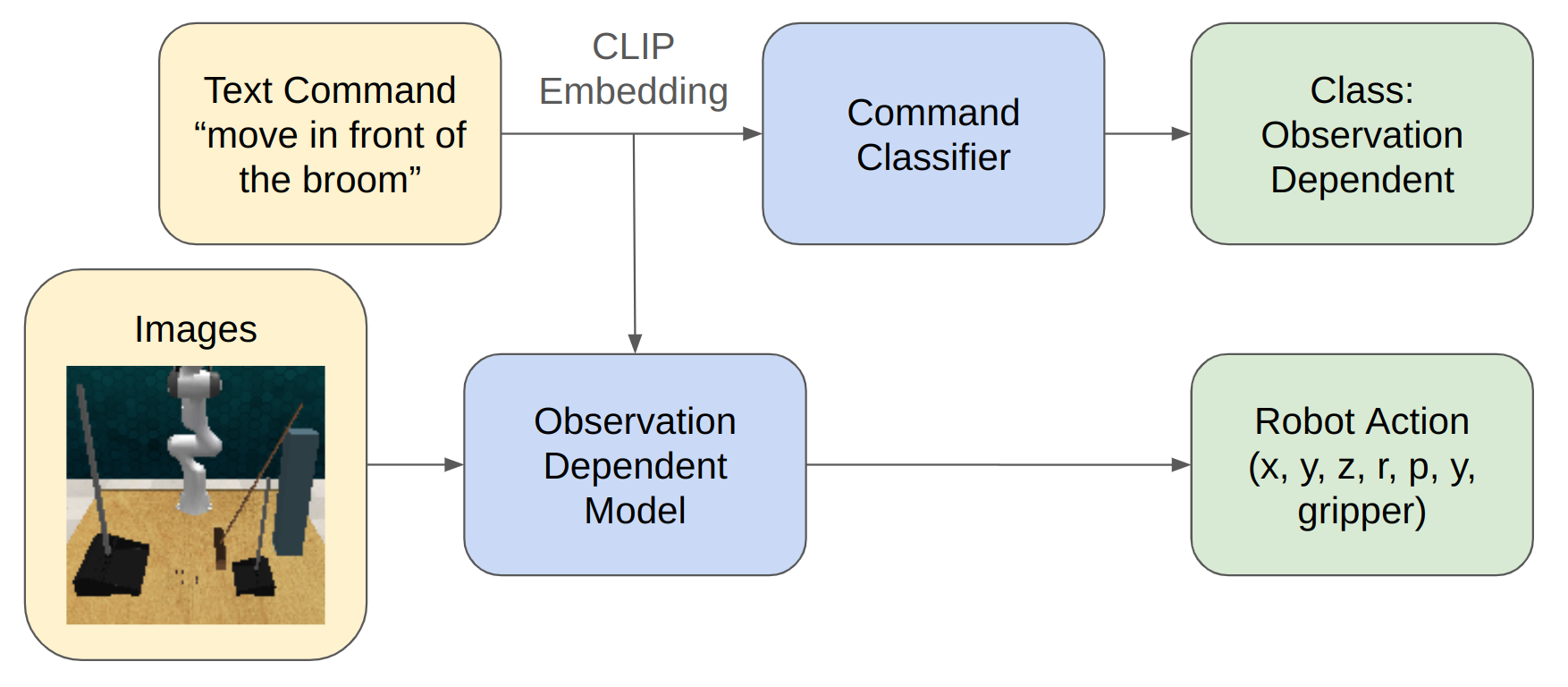}
    \caption{If the command classifier determines a command is observation-dependent, the observation-dependent model uses the text embedding and the current image observations to output a robot action, as shown above.
    }
    \end{subfigure}
\caption{Our architecture includes a command classifier which determines whether to run an observation-dependent or observation-independent model. The \textit{factory model} and \textit{home model} include both models. The observation-dependent model is fine-tuned in the \textit{home model}.}
\label{fig:architecture}
\end{figure}

\begin{figure*}
\centering
\includegraphics[width=0.95\textwidth]{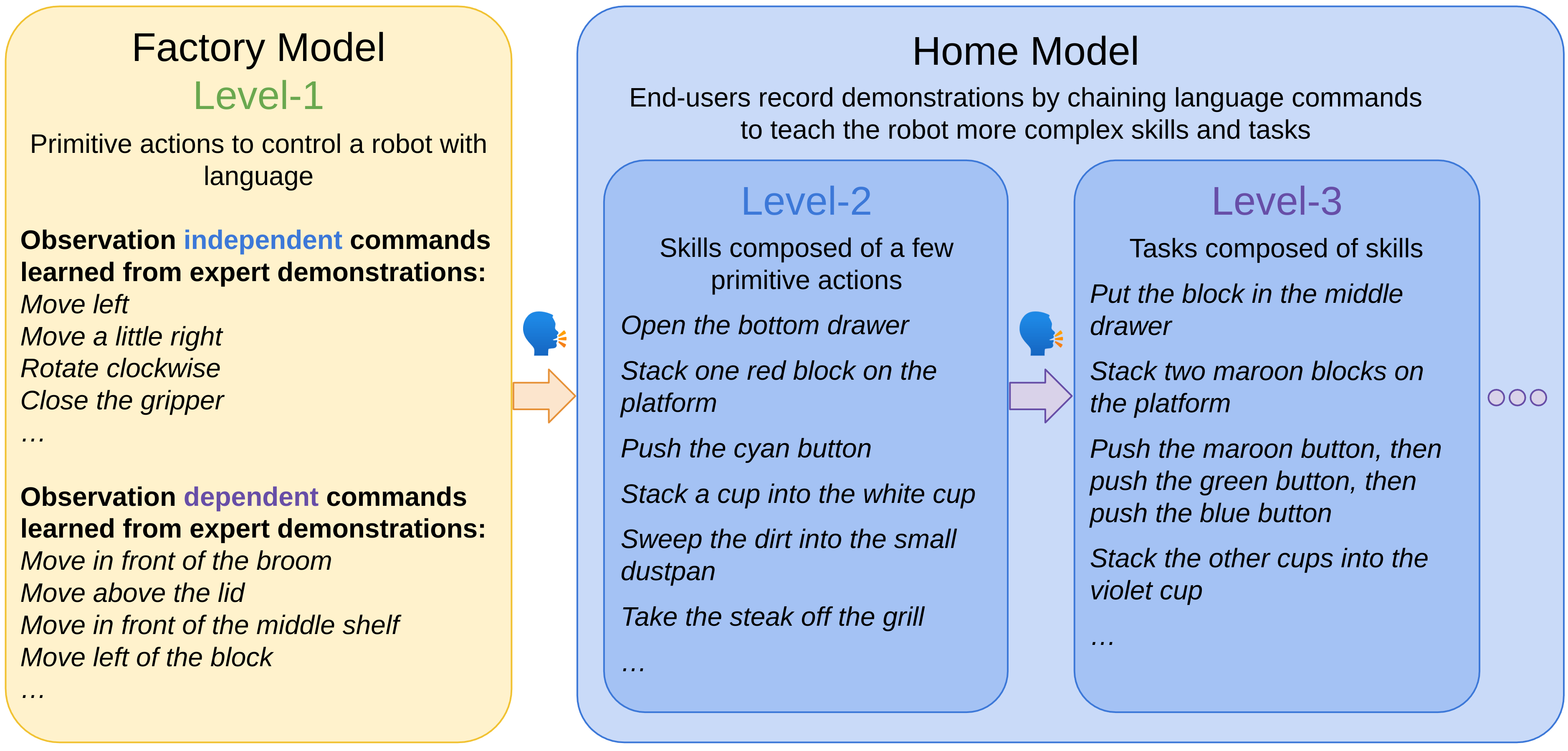}
  \caption{The Level-1 \textit{factory model} is trained on scripted demonstrations to perform primitive actions from language commands. An end user trains the robot to perform Level-2 skills in their home by using the Level-1 action commands to collect demonstrations of desired skills. They can then train Level-3 tasks by utilizing Level-1 action commands and Level-2 skill commands to collect demonstrations of desired tasks. These demonstrations collected by the end user only use natural language; no programming or special hardware is required. Different end users may choose to train different skills and tasks according to their needs.}
  \label{fig:home_model}
\end{figure*}

Our network architecture consists of an observation-dependent, and an observation-independent model, as shown in Figure~\ref{fig:architecture}. The command classifier determines which model to use for a given text command. Both models take in a text instruction and output a robot action. The observation-dependent model takes in RGB-D images in addition to the text instruction. This architecture applies to the \textit{factory model} and the \textit{home model} illustrated in Figure~\ref{fig:home_model}. The \textit{home model} is a \textit{factory model} trained on more observation-dependent commands; there is no change to the model architecture.

The observation-independent model is a multi-layer perceptron that regresses $\Delta$x, $\Delta$y, $\Delta$z, $\Delta$roll, $\Delta$pitch, $\Delta$yaw, and gripper state (open or close) from a CLIP~\cite{radford21clip} embedding of the text instruction. The model also takes the previous output as input to maintain the previous gripper state. We train using a set of labeled commands. For example, ``move left" is a 10cm move in the negative x direction, and ``rotate clockwise" is a 90-degree roll. Some of these commands and labels are shown in Appendix~Table~\ref{table_commands}.

The observation-dependent model is PerAct~\cite{shridhar2022peract} with minor modifications. This model has a preprocessing stage, which converts the RGB-D images into a voxel representation. We elected to reduce the voxel dimensions to 50$\times$50$\times$50 from 100$\times$100$\times$100 to save computation, enabling us to run more experiments in a shorter time. Note that our architecture is modular and hence any advancements in feature embedding and efficient observation-dependent models can be integrated easily. 

\subsection{Model Levels}
The Level-1 model (aka \textit{factory model}) is trained on primitive motions demonstrated by a scripted expert. In our experiments, these are scripted trajectories from the RLBench environment. The primitive motions include both observation-independent commands (e.g. `Move a little right'), and observation-dependent commands (e.g. `Move above the block'). Level-2 and Level-3 are skill and task models, respectively, trained by end users via language-commanded demonstrations. The skills model includes things like picking and placing an object, or pushing an object to a specified location. The tasks model includes things that require repeating a skill or combining multiple skills. Tasks include repeating a pick and place to stack multiple objects or pulling a drawer open, then putting a block in the drawer. Refer to the Figure~\ref{fig:home_model} where the Level-1 model is the \textit{factory model} and Level-2 and beyond are \textit{home models}.

\subsection{Environments}
We selected 14 RLBench tasks, which are a subset of the 18 tasks evaluated in PerAct. Each voxel in our model is 2cm wide, which makes high precision tasks more difficult. Therefore, we eliminated the tasks of screwing in a light bulb, sorting shapes in a shape sorter, placing a ring on a peg, and hanging mugs on a mug tree because they require very high precision. To avoid confusion, we refer to the 14 RLBench tasks as environments. Some of the RLBench tasks are actually skills according to our definitions. Each environment is used to train Level-1 motions, and Level-2 skills. Some environments are also used to train Level-3 tasks.

\subsection{Level-1 Factory Model}
The basis of our approach is training a Level-1 model on demonstrations of primitive motions. The demonstrations for Level-1 all consist of a single robot motion. We created primitive action demonstrations for each RLBench environment. For example, the \textit{open drawer} environment has 3 primitive actions: ``move in front of the top handle," ``move in front of the middle handle," and ``move in front of the bottom handle." The \textit{factory model} is trained with 1400 scripted demonstrations covering primitive motions across all the environments. In practice, these scripted demonstrations will be replaced by expert demonstrations in the factory setting. 

\subsection{Collecting Demonstrations with Language}
\begin{figure}
\includegraphics[width=0.5\textwidth]{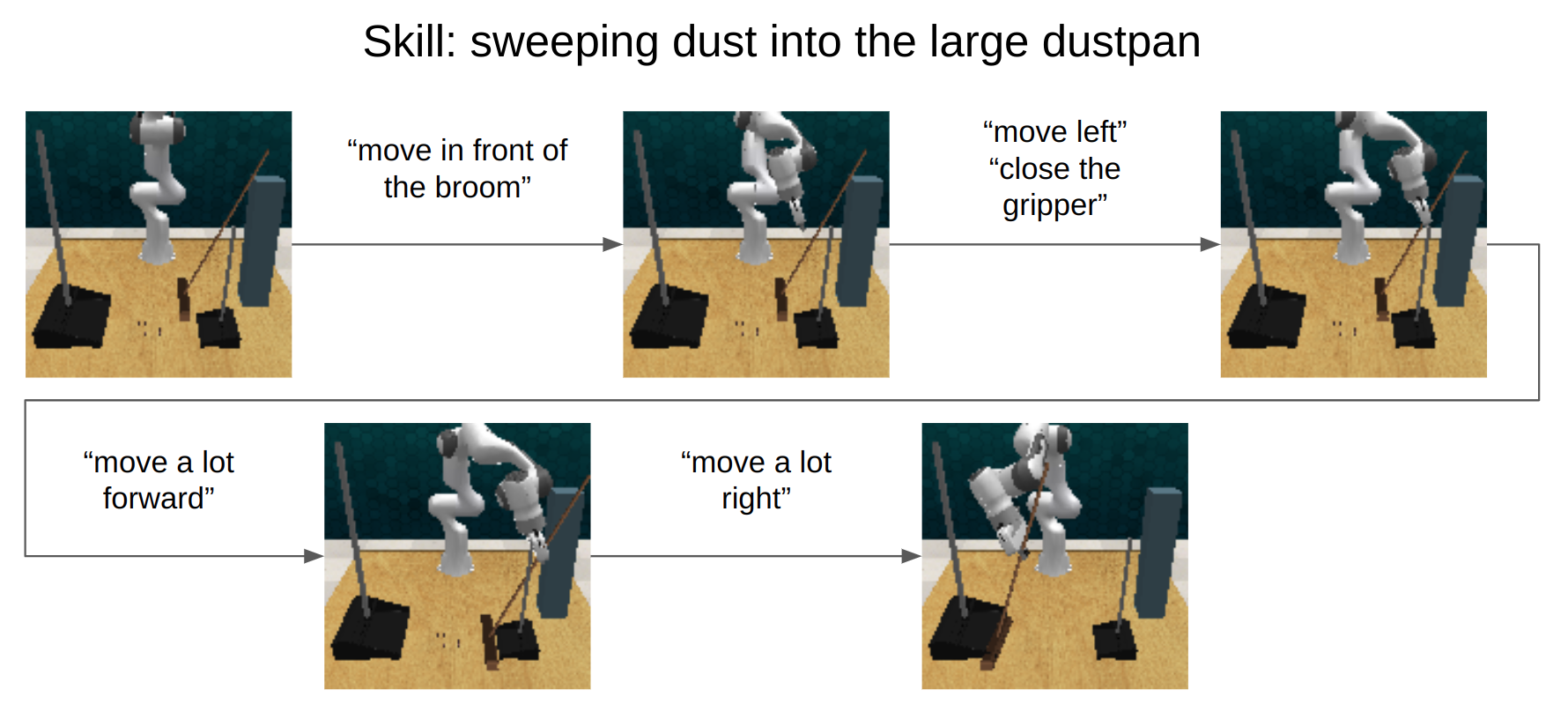}
  \caption{Primitive motion  (Level-1) commands are used to collect a skill (Level-2) demonstration of sweeping dust into the large dustpan.}
  \label{fig_demo_collection}
\end{figure}

Once the \textit{factory model} is trained, we no longer need any scripted demonstrations. Demonstrations for more complex skills and tasks are collected using only language instructions typed by the user. Figure \ref{fig_demo_collection} shows a demonstration of the Level-1 language commands used to sweep dirt into a dustpan. The keyframes are labeled by the user during the language-driven demonstration so it can be used to train this skill.

\subsection{Home Model Training}
Once the user collects a few demonstrations of a skill or task they desire the robot to learn, they can have the \textit{factory model} fine-tune these demonstrations to create their \textit{home models} (in our case Level-2 and Level-3). Different users may care more about different tasks and only fine-tune for a few of the many possible tasks the robot could learn.

%% file: sections/03_experimental_setup.tex
\section{Experiments \& Results}
In this section, we discuss the details of our experimental setups, and the results of each experiment. For all experiments, model weights are saved after every 5000 steps of training. Every saved weight is evaluated on 25 unseen rollouts of each skill or task the model was trained on. The evaluation episodes are seeded so they initialize repeatably. The best weight from the evaluation is then tested on another set of rollouts. The test episodes are seeded to ensure they are different than the evaluation episodes.

The first experiment determines how well the \textit{factory model} learns to complete primitive actions. We follow that with a language augmentation experiment to see if we can train a \textit{factory model} that is robust to paraphrased inputs.

Next, we compare Level-2 \textit{home models} with a baseline Level-2 model. The \textit{home models} include a multi-skill model, and single-skill models. As part of our Level-2 analysis, we test the Level-2 multi-skill model with Level-1 action commands. We want to make sure the model doesn't forget earlier levels. 

We then compare Level-3 \textit{home models} with a baseline Level-3 model. The \textit{home models} include a multi-skill model, and single-skill models.

Last, we experiment with a VLM to see if we can avoid having a human collect demonstrations. We try using the VLM to break down Level-2 skills into Level-1 primitive action commands, and to break down Level-3 tasks into Level-2 skill commands.

\subsection{Learning Level-1 Primitive Actions}
For primitive actions, success is determined by the robot end effector reaching within 2.5cm of the target position. The model is given up to 5 action predictions to reach the target position. On average, the robot reaches the correct location 87\% of the time. This Level-1 model is a strong foundation for collecting demonstrations for higher-level models via language. 

\subsection{Language Augmentation}
We can't expect end users to give commands identical to those we used. In order to test variations in language, we created 6 total paraphrases of the Level-1 action instructions. We refer to these as variations $i={0, 1, 2, 3, 4, 5}$. We use a Level-1 model that was only trained on variation $i=0$ as the default model. We compare it to a Level-1 model that was trained on variations $i=0, 1, 2, 3$. Variations $i=4, 5$ were not seen by either model. Variation $i=4$ uses a novel combination of words, but all the individual words were seen in variations $i=0, 1, 2, 3$. Variation $i=5$ includes at least one word that was not seen in any of the other variations.

The augmented model shows an 8\% improvement in success rate compared to the default model on both test variations, shown in Table \ref{table_lang_aug}. The success rate is averaged across all the motion commands in all 14 RLBench environments. This shows a model trained on more language variations is more likely to succeed when given paraphrased language commands not seen in training.

\begin{table}[H]
\centering
\begin{tabular}{|c|c|c|}
    \hline
    Model & Variation 4 & Variation 5 \\
    \hline
    Default & 76 & 75 \\
    \hline
    Augmented & 84 & 83 \\
    \hline
\end{tabular}
\caption{Comparison between a default Level-1 model trained with no language variation, and an augmented Level-1 model trained with 4 language variations. Variation 4 and variation 5 were not seen by either model during training.}
\label{table_lang_aug}
\end{table}

\subsection{Baseline Models}
We create baseline models for Level-2 and Level-3 using a more traditional approach. We generate 10 demos for each skill using scripted waypoints in RLBench. Then we train a model on the scripted demos for 100,000 steps. The key differences are that our method trains the model with Level-1 demos first, and our method uses Level-2 demos collected with language instead of using scripted waypoints. The same is true for the Level-3 baseline; it is trained on 10 demos for each task using scripted waypoints.

\subsection{Learning Level-2 Skills from Level-1 Actions}
From our 14 RLBench environments, we define 14 Level-2 skills and 4 Level-3 tasks. The skills are listed in Table~\ref{table:level2Multi}. The tasks build on the skills. For example, \textit{put in drawer} includes the skills of \textit{open drawer}, and \textit{put in drawer}.

We train two types of Level-2 models - a) multi-skill model and b) single-skill model. We envision that an end user would want to train for a subset of the possible skills, so we experiment to understand the performance when all skills are learned at the same time vs. single skills. We use 10 demos for each skill to train these models. The training starts with the Level-1 weights and fine-tunes for 100,000 steps. During the fine-tuning, each batch samples three times from the Level-1 dataset and once from the Level-2 dataset. This sampling prevents the model from forgetting Level-1 actions. We find that the Level-2 model achieves an average success rate 7\% higher than the Level-1 model when tested on Level-1 actions, as shown in Table \ref{table_motions_success}, meaning it improves rather than forgets.

\begin{table}[h!]
\centering
    \begin{tabular}{|c|c|}
    \hline
    Model & Average Success Rate on Level-1 Actions \\
    \hline
    Level-1 & 87 \\
    \hline
    Level-2 & 94 \\
    \hline
    \end{tabular}
    \caption{We test the Level-2 model on Level-1 actions to determine whether it is forgetting Level-1 commands. Rather than forgetting, it shows improvement over the Level-1 model.}
    \label{table_motions_success}
\end{table}

The multi-skill and single-skill models are evaluated every 5000 steps on 25 episodes of the skills. The models that perform best on the validation episodes are used for testing, once again on 25 episodes per skill. Our method produces a multi-skill model that achieves an average success rate of 39\% compared to the baseline of 23\%, as seen in Table~\ref{table:level2Multi}. We believe many important features for learning a skill are learned from the Level-1 actions model. In the open drawer skill, the Level-1 model already learned the concepts of top, middle, and bottom from Level-1. In the push buttons skill, the Level-1 model already learned the 18 button color variations. Table~\ref{table_single_lv2} shows that performance is higher when a model is trained on a single skill instead of many skills, demonstrating that models trained for specific use cases are better than more generalist models.

\begin{table*}
\centering
\begin{tabular}{|c|c|c|c|c|c|c|c|}
    \hline
    Model & Average & Open Drawer & Slide Block & Sweep to Dustpan & Meat Off Grill & Turn Tap & Put in Drawer Lv2 \\
    \hline
    Ours 5 demos & 30$\pm$2 & \textbf{79$\pm$6} & 24$\pm$8 & 33$\pm$22 & 5$\pm$2 & \textbf{52$\pm$8} & 60$\pm$16 \\
    \hline
    Ours 10 demos & \textbf{39$\pm$2} & \textbf{79$\pm$6} & 40$\pm$4 & \textbf{68$\pm$11} & 15$\pm$15 & 51$\pm$14 & \textbf{84$\pm$7} \\
    \hline
    Baseline 10 demos & 23$\pm$3 & 49$\pm$9 & \textbf{43$\pm$5} & 23$\pm$21 & \textbf{41$\pm$2} & 8$\pm$7 & 44$\pm$31 \\
    \hline
    \hline
    Close Jar & Drag Stick & Stack Blocks Lv2 & Put in Safe & Place Wine & Put in Cupboard & Push Buttons Lv2 & Stack Cups Lv2 \\
    \hline
    13$\pm$2 & \textbf{56$\pm$7} & 5$\pm$5 & 3$\pm$2 & 7$\pm$2 & 0 & 73$\pm$6 & 8$\pm$4 \\
    \hline
    \textbf{28$\pm$7} & 51$\pm$27 & \textbf{9$\pm$5} & 16$\pm$7 & 8$\pm$7 & 0 & \textbf{79$\pm$10} & \textbf{15$\pm$2} \\
    \hline
    3$\pm$2 & 24$\pm$7 & 7$\pm$6 & \textbf{21$\pm$6} & \textbf{23$\pm$8} & 0 & 24$\pm$11 & 9$\pm$5 \\
    \hline
\end{tabular}
\caption{Success rates for the multi-skill Level-2 model compared to a baseline. Success rates are the mean and standard deviation from tests on 3 models initialized randomly before training.}
\label{table:level2Multi}
\end{table*}

\begin{table*}
\centering
\begin{tabular}{|c|c|c|c|c|c|c|c|c|}
    \hline
    Average & Open Drawer & Slide Block & Sweep to Dustpan & Meat Off Grill & Turn Tap & Put in Drawer Lv2 & Close Jar\\
    \hline
    50 & 84 & 60 & 80 & 36 & 68 & 92 & 32\\
    \hline
    Drag Stick & Stack Blocks Lv2 & Put in Safe & Place Wine & Put in Cupboard & Push Buttons Lv2 & Stack Cups Lv2 &  \\
    \hline
    80 & 4 & 28 & 32 & 0 & 80 & 20 & \\
    \hline
\end{tabular}
\caption{Success rates for Level-2 models fine-tuned for a single skill}
\label{table_single_lv2}
\end{table*}

\subsection{Learning Level-3 Tasks from Level-1 \& 2}
To learn Level-3 tasks, we fine-tune for 100,000 steps with Level-3 demos on the best Level-2 models. Fine-tuning here means each batch takes 2 samples from Level-1, 1 sample from Level-2, and 1 sample from Level-3. We train single-task and multi-task models, and the results are shown in Tables~\ref{table_multi_lv3} and \ref{table_single_lv3}. The baseline multi-task model achieves an average success rate of 10\%, whereas our multi-task model achieves 23\% using the same number of Level-3 demos. Even our multi-task model trained on only 5 Level-3 demos achieves a 7\% better success rate than the baseline. Curiously, our model trained on 5 demos appears to beat the more thoroughly trained models on the \textit{stack blocks} task, but we attribute that result to chance and deem it statistically insignificant.

\begin{table*}[th!]
\centering
\begin{tabular}{|c|c|c|c|c|c|}
    \hline
    Model & Average & Put in Drawer & Stack Blocks & Push Buttons & Stack Cups \\
    \hline
    Ours 5 demos & 17$\pm$2 & 28$\pm$11 & \textbf{3$\pm$5} & 39$\pm$6 & 0 \\
    \hline
    Ours 10 demos & \textbf{23$\pm$6} & \textbf{40$\pm$8} & 0 & \textbf{53$\pm$17} & 0 \\
    \hline
    Baseline 10 demos & 10$\pm$2 & 0 & 0 & 39$\pm$9 & 0 \\
    \hline
\end{tabular}
\caption{Success rates for Level-3 tasks with multi-task models. Success rates are the mean and standard deviation from tests on 3 models initialized randomly before training.}
\label{table_multi_lv3}
\end{table*}

\begin{table*}
\centering
\begin{tabular}{|c|c|c|c|c|c|}
    \hline
    Demos & Average & Put in Drawer & Stack Blocks & Push Buttons & Stack Cups \\
    \hline
    5 demos & 25 & 40 & 4 & 56 & 0 \\
    \hline
    10 demos & 24 & 36 & 0 & 56 & 4 \\
    \hline
\end{tabular}
\caption{Success rates for Level-3 models fine-tuned for a single task}
\label{table_single_lv3}
\end{table*}

\subsection{Using Large Vision-Language Models} \label{experiments-vlm}
\begin{figure}
\centering
\includegraphics[width=0.5\textwidth]{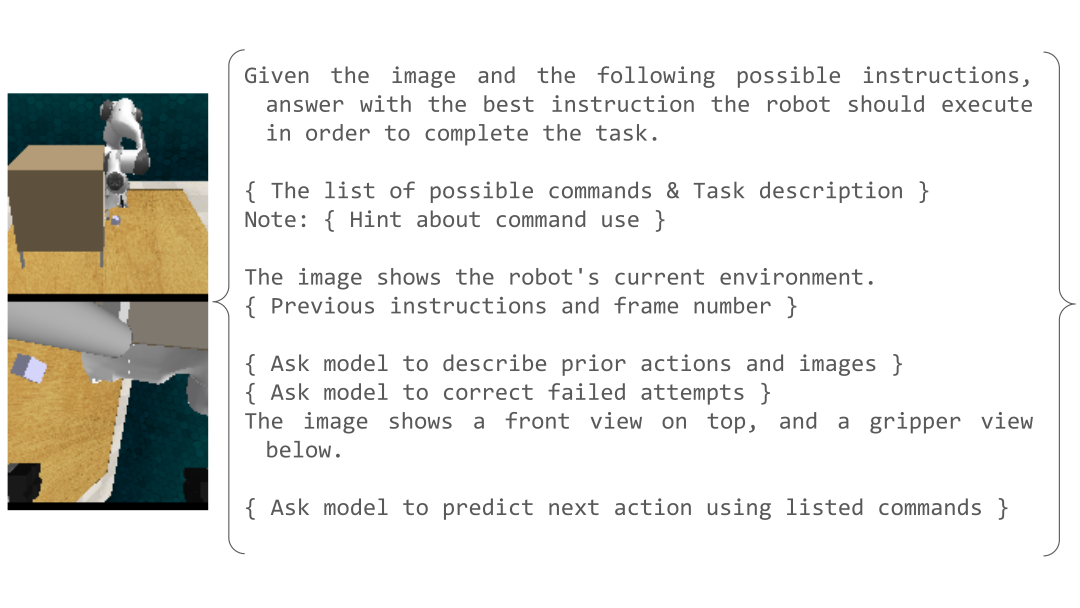}
\vspace{-10mm}
\caption{The prompt template shown above is used to query the VLM for the next actions. Every action proposed is executed by the policy for 8 steps. The images are updated after every execution.}
\label{fig:vlm_prompt}
\end{figure}

Inspired by PaLM-E \cite{driess2023palm}, which demonstrated the application of pre-trained LLM-based VLMs to control downstream primitive policies, we explore utilizing VLM-generated lower-level instructions to complete higher-level skills or tasks. We use Bard (\textit{version 2023.10.30}) as it is the most advanced VLM accessible at the time of this work, but GPT4 \cite{openai2023gpt4} or other VLMs could be substituted. We prompt Google's Bard \cite{anil2023palm} with a list of possible actions and an image containing a front view and a gripper view, as shown in Figure~\ref{fig:vlm_prompt}. We prompt the VLM to output a description of the image, a list of past executed actions, and whether or not the task has been successfully executed. The VLM is then asked to state the feasibility of each potential next action and choose the best one. This idea is similar to the idea of inner monologue \cite{huang2022inner}, where the prompt draws attention to the image and previous actions. The list of possible actions is reduced to only the relevant actions for a given task. We observe that an off-the-shelf VLM model which has not been trained on any manipulation tasks fails at lower-level grounded reasoning, but excels in higher-level task planning.

In the following sections, we picture the VLM as a reasoning tool and our Level-1 or Level-2 model as a policy function that the VLM can utilize.

\textbf{VLM Reasoning over Level-1 Policy: }
We first provide the VLM with a \textit{factory model} (Level-1) and observe that the VLM fails at lower-level grounded reasoning and thus struggles to perform zero-shot tasks using the \textit{factory model}.
In this experiment, the VLM is given a list of Level-1 commands to use to complete the skill. We observe that the VLM fails to perform 3D spatial reasoning and provides poor justification as to why it elected to perform an action. We hypothesize that the VLM reasons at a high level, which is corroborated by the way it describes a scene when prompted. Similarly, the VLM was not trained to comprehend the robot's state and tends to predict that the robot is carrying an object even when the gripper is open. This demonstrates the need for the user to extend the \textit{factory model} to a Level-2 skill model before attempting to leverage a VLM.

\textbf{VLM Reasoning over Level-2 Policy: } We provide the VLM with a Level-2 multi-skill model and observe that it is capable of utilizing it to achieve some success in zero-shot task execution. The VLM is able to complete the \textit{put in drawer} and \textit{push buttons} tasks. We do not attempt \textit{stack blocks} or \textit{stack cups} since the provided Level-2 policy already has low success rates on the prerequisite skills (9\% and 15\%, respectively). 

\begin{table}
\centering
\begin{tabular}{|c|c|c|c|}
    \hline
    Model & Average & Put in Drawer & Push Buttons \\
    \hline
    Best L3 & 48 & 40 & 56 \\
    \hline
    VLM & 57.5 & 25 & 90 \\
    \hline
\end{tabular}
\caption{Success rates of Level-3 single-task models compared to the VLM using a Level-2 multi-skill policy}
\label{table:vlm}
\end{table}

On average, the VLM performs better than the trained Level-3 models, as shown in Table~\ref{table:vlm}. This result suggests that the VLM can reason well over a high-level task given a prompt. The performance was better on \textit{push buttons} and worse on \textit{put in drawer}. We observe that a primary reason for the VLM's failure on \textit{put in drawer} is that the Level-2 policy fails to open the drawer, and the VLM fails to recognize that and does not retry.

%% file: sections/04_conclusion.tex
\section{Conclusion}
We present a framework that is designed for end users to train a robot to perform a variety of skills and tasks. By providing a \textit{factory model} capable of following language instructions for primitive actions, we show longer-horizon demonstrations can be collected using only natural language. Our hierarchical training method produces \textit{home models} that achieve a 1.7x improvement on Level-2 skills, and a 2.3x improvement on Level-3 tasks compared to the baseline method. We hope that the findings of this work can encourage research toward end-user directed robot training methods. 

We attempt to replace the human in the loop with a VLM, but find the VLM falls short on low-level reasoning. The VLM lacks the spatial reasoning abilities to make fine adjustments to the robot position to complete Level-2 skills using Level-1 primitive action commands. The VLM shows better results using Level-2 skill commands to complete Level-3 tasks. These results show potential benefits of coupling our system with a VLM.

One limitation of this work is the limited selection of training environments. The \textit{factory model} must be trained with data from a wide variety of tasks and environments to function in homes. Increasing the variety of training environments is a potential future work.

%% file: sections/appendix.tex
\subsection{Training Data}
This section provides more information on the training of the Level-1 \textit{factory model}. Table~\ref{table_commands} shows examples of labels for observation-independent commands. Table \ref{table_level1} gives the success rate breakdown of the primitive actions in each RLBench environment. The motions for the \textit{turn tap} environment have a low success rate because the robot is trying to move directly to a grasp position. We train a pre-grasp position for most environments, but in the \textit{turn tap} environment, the tap handles are at angles difficult to describe with language. The actions for the \textit{put in cupboard} environment also have a low success rate. This is because the grocery items are not easy to distinguish in our low-resolution voxel space. Some of the box shaped or cylinder shaped items are easily confused with each other.

\begin{table}[h]
    \centering
    \tabcolsep=0.1cm
    \begin{tabular}{|c|c|c|c|c|c|c|c|}
        \hline
        Command & $\Delta$x & $\Delta$y & $\Delta$z & $\Delta$roll & $\Delta$pitch & $\Delta$yaw & gripper \\
        \hline
        move left & -10 & 0 & 0 & 0 & 0 & 0 & 1 \\
        \hline
        move a little left & -5 & 0 & 0 & 0 & 0 & 0 & 1 \\
        \hline
        move a lot left & -20 &	0 &	0 &	0 &	0 &	0 &	1 \\
        \hline
        move a tiny bit left & -1 &	0 &	0 &	0 &	0 &	0 &	1 \\
        \hline
        move forward & 0 & 10 &	0 &	0 &	0 &	0 &	1 \\
        \hline
        move up & 0 & 0 & 10 & 0 & 0 & 0 & 1 \\
        \hline
        move backward and down & 0 & -10 & -10 & 0 & 0 & 0 & 1 \\
        \hline
        rotate clockwise & 0 & 0 & 0 & 90 &	0 &	0 &	1 \\
        \hline
        turn left & 0 &	0 &	0 &	0 &	0 & -90 & 1 \\
        \hline
        turn up & 0 &	0 &	0 &	0 &	-90 &	0 &	1 \\
        \hline
        close the gripper &	0 &	0 &	0 &	0 &	0 & 0 &	0 \\
        \hline
    \end{tabular}
    \caption{Selected examples of commands and labels used to train the observation-independent model}
    \label{table_commands}
\end{table}

\renewcommand\tabularxcolumn[1]{m{#1}}
\begin{table}[h!]
    \centering
    \begin{tabularx}{0.5\textwidth}{|l|X|l|}
        \hline
        Environment & Motions & Success Rate \\
        \hline
        Open Drawer & move in front of the \{top, middle, bottom\} handle & 96 \\
        \hline
        Slide Block & move \{in front of, behind, left of, right of\} the block & 100 \\
        \hline
        Sweep to Dustpan & move in front of the broom & 100 \\
        \hline
        Meat Off Grill & move above the \{steak, chicken\} & 100 \\
        \hline
        Turn Tap & move to the \{left, right\} tap & 36 \\
        \hline 
        Put in Drawer & move above the block & 92 \\
        \hline
        Close Jar & move above the \{color\} jar, \newline move above the lid & 92 \\
        \hline
        Drag Stick & move above the stick & 100 \\
        \hline
        Put in Safe & move above the money, \newline move in front of the \{top, middle, bottom\} shelf & 96 \\
        \hline
        Place Wine & move in front of the wine bottle, \newline move in front of the \{near side, middle, far side\} of the rack & 96 \\
        \hline
        Put in Cupboard & move above the \{item\}, \newline move in front of the cupboard & 48 \\
        \hline
        Push Buttons & move above the \{color\} button & 92 \\
        \hline
        Stack Cups & move above the left edge of the \{color\} cup & 88 \\
        \hline
    \end{tabularx}
    \caption{Level-1 actions model success rates for 25 test episodes in each environment}
    \label{table_level1}
\end{table}

\newpage
\subsection{VLM Experiment Details}

\begin{figure}
    \centering
    \includegraphics[width=0.5\textwidth]{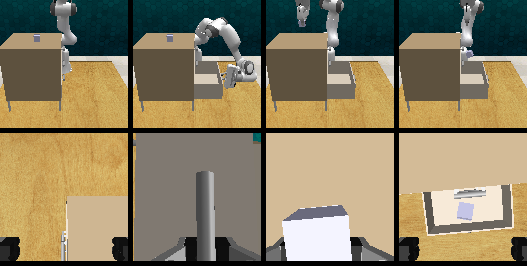}
    \caption{VLM success for \textit{Put in Drawer}. The VLM correctly predicts the sequence of commands (left to right): \textit{open the bottom drawer}, \textit{put the block in the bottom drawer}, \textit{move a lot left} (irrelevant since the task is complete)}
    \label{fig:vlm_put_in_drawer_success}
\end{figure}

\begin{figure}
    \centering
    \includegraphics[width=0.5\textwidth]{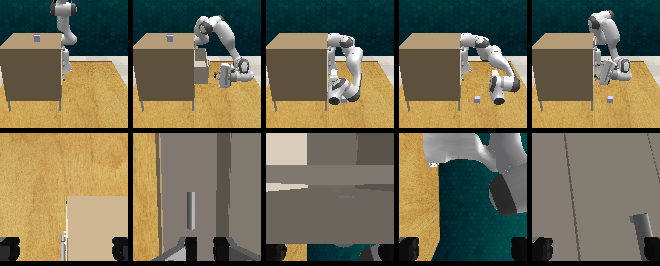}
    \captionof{figure}{VLM failure for \textit{Put in Drawer}. The VLM correctly predicts the sequence of commands, but the Level-2 model fails to execute them correctly. Predicted commands (left to right): \textit{open the middle drawer}, \textit{put the block in the middle drawer}, \textit{move a lot left}, \textit{put the block in the middle drawer}.
    \label{fig:vlm_put_in_drawer_fail}}
\end{figure}

\begin{figure*}[t!]
    \centering
    \includegraphics[width=.7\textwidth]{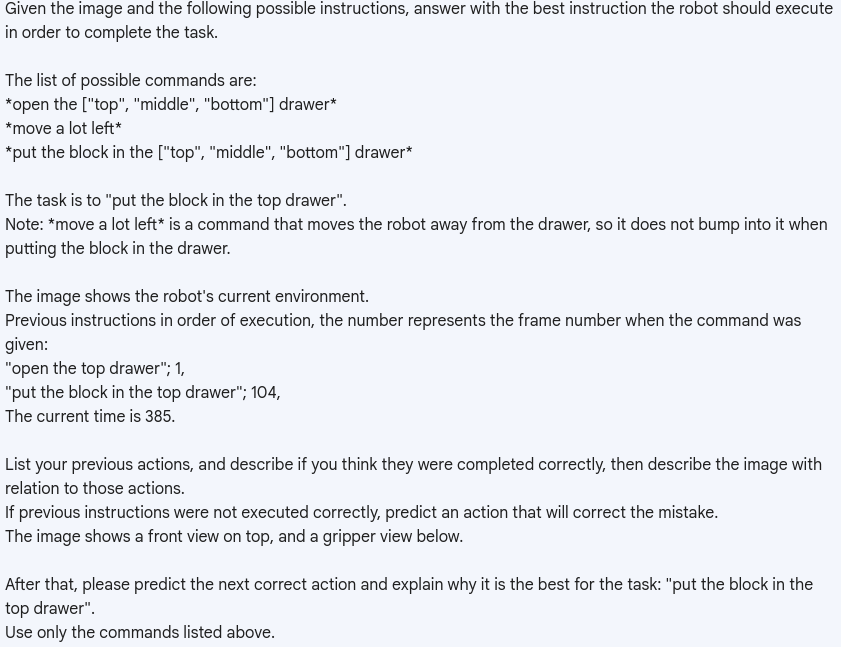}
    \includegraphics[width=0.25\textwidth]{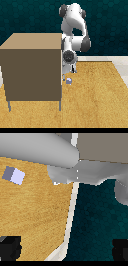}
    \caption{Text prompt for VLM tasks. The VLM is given a prompt with the list of possible actions, and the history of previous actions, along with a task in each prompt. Each prompt includes the current images from the front view and gripper view. The output of the VLM is sent to the Level-2 skills model. The Level-2 skills model is allowed to run for 8 steps, then the VLM is prompted with the updated images.}
    \vspace*{12.5cm}
    \label{fig:vlm_prompt_full}
\end{figure*}

This section provides examples and details of our VLM experiments. These experiments involved prompting the VLM with task objective details, possible actions to perform, a list of previously performed actions, and a description of the requested output. Alongside the description, images showing the current state of the robot are also provided. Figure \ref{fig:vlm_prompt_full} shows an example of a complete prompt used in our VLM experiment. Figure \ref{fig:vlm_put_in_drawer_success} shows a successful use of the VLM whereas Figure \ref{fig:vlm_put_in_drawer_fail} shows a failure case. 

